# Russia-Ukraine war: Modeling and Clustering the Sentiments Trends of Various Countries

Hamed Vahdat-Nejad*, Mohammad Ghasem Akbari**, Fatemeh Salmani*, Faezeh Azizi*,

Hamid-Reza Nili-Sani**

* PerLab, Faculty of Electrical and Computer Engineering, University of Birjand, Birjand, Iran

**Department of Statistics, University of Birjand, Birjand, Iran

Email: vahdatnejad@birjand.ac.ir, g_z_akbari@birjand.ac.ir, salmani_fatemeh98@birjand.ac.ir,

faezeh.azizi1995@birjand.ac.ir, hnilisani@birjand.ac.ir

**Abstract**

With Twitter's growth and popularity, a huge number of views are shared by users on various topics, making this platform a valuable information source on various political, social, and economic issues. This paper investigates English tweets on the Russia-Ukraine war to analyze trends reflecting users' opinions and sentiments regarding the conflict. The tweets' positive and negative sentiments are analyzed using a BERT-based model, and the time series associated with the frequency of positive and negative tweets for various countries is calculated. Then, we propose a method based on the neighborhood average for modeling and clustering the time series of countries. The clustering results provide valuable insight into public opinion regarding this conflict. Among other things, we can mention the similar thoughts of users from the United States, Canada, the United Kingdom, and most Western European countries versus the shared views of Eastern European, Scandinavian, Asian, and South American nations toward the conflict.

**1. Introduction**

Social networks are currently one of the most significant platforms for population information extraction. Users of social networks share innumerable posts in many domains, including social, economic, and political ones, that reflect their perspectives and sentiments regarding global events and incidents (Crannell, Clark, Jones, James, & Moore, 2016). As a social network, Twitter is a key source of users' sentiments and perspectives. Users' views can be analyzed using natural language processing and data retrieval.

Natural language processing (NLP) is "focused on the design and analysis of computational algorithms and representations for processing natural human language" (Eisenstein, 2018). It aims to provide "new computational capabilities around human language" (Eisenstein, 2018). Indeed, we can extract information from a mass of texts using NLP techniques.

Sentiment analysis is a promising technique used in natural language processing to extract emotional states and subjective information from a text (Zad, Heidari, Jones, & Uzuner, 2021). It allows for the extraction of sentiments from users' tweets regarding a variety of events and issues. The outbreak of COVID-19, for instance, was a globally important event about which many users have expressed their opinions. Alongside this, there have been sentiment analyses of tweets related to COVID-19 (Vahdat-Nejad et al., 2022), the economy during COVID-19 (Salmani, Vahdat-Nejad, & Hajiabadi, 2021), and education during the epidemic of COVID-19 (Jamalian, Vahdat-Nejad, & Hajiabadi, 2022), among others. Other events, including wars and conflicts, have also prompted users to post and analyze tweets. For instance, the sentiment analysis of Turkish and English tweets about Syrian refugees has revealed a predominantly positive tone in Turkish tweets but a negative tone in English tweets (Öztürk & Ayvaz, 2018).

Similarly, an analysis of opinions regarding recent events in Afghanistan has revealed that tweets containing negative terms such as *terrorist*, *attack*, *destroy*, and *violence* have dominated

(Aggarwal, Khan, & Kakkar, 2022). Furthermore, the ratio of negative to positive tweets was identical across all eight investigated countries, indicating that these nations shared similar sentiments regarding the recent events in Afghanistan (Aggarwal et al., 2022). Likewise, the sentiment analysis of tweets about the Syrian chemical war reveals that users negatively view the Syrian chemical attack (Bashir et al., 2021).

The Russia-Ukraine war has probably been the most prominent event of 2022. Recently, the main topics discussed on the Chinese Weibo social network regarding this conflict have been extracted (Chen et al., 2022). To our best knowledge, no study has analyzed tweets regarding this war to investigate and categorize the attitude of various countries. To this end, this research collects and analyzes tweets on this conflict to cluster various countries according to their population's attitudes. For this purpose, 140,000 tweets posted in the first month of the war (March 2022) were collected using the keyword *Ukraine*. In order to analyze the sentiments of tweets, the language-based RoBERTa (Liu et al., 2019) model is employed, which has superior accuracy and performance compared to similar models (Briskilal & Subalalitha, 2022). Following this, the time series of the frequency of positive and negative sentiments are calculated and modeled for several countries that have a sufficient number of English tweets. Next, the time series models of the countries are clustered based on the neighborhood average method. Results indicate that countries such as the United States, England, Canada, and those of Western Europe are in one cluster, while Eastern Europe, Scandinavian, South American, and Asian countries are in another. In addition, Ukraine was placed alone in one cluster, indicating a divergent trend in public opinion between this country and the rest of the world in the first month of the conflict. The main contributions of the paper are as follows:

- To the best of our knowledge, it is the first work investigating and clustering the attitudes of nations on Twitter regarding the Ukraine war.

- It proposes a new method for modeling and clustering the time series of countries' sentiments.

The article's remaining sections are as follows: Section 2 reviews previous research. Section 3 describes the proposed methodology. Implementation and results are described in Section 4. Lastly, the fifth section summarizes the conclusions and limitations.

## 2- Related work

As a result of their growth and popularity, social networks have become an important and valuable information source on users' attitudes in various fields. Techniques such as big data and natural language processing are required to analyze this unstructured textual information of social network users. Twitter is one of the most popular social networks on which users share their thoughts and feelings about current events and issues (Mathur, Kubde, & Vaidya, 2020). Numerous analyses of user opinions have been conducted in a variety of fields, including politics (Ansari, Aziz, Siddiqui, Mehra, & Singh, 2020), stock return forecasting (Sul, Dennis, & Yuan, 2017), sustainable energy (Corbett & Savarimuthu, 2022), and tourism (Abbasi-Moud, Vahdat-Nejad, & Mansoor, 2019; Asani, Vahdatnejad, Hosseinabadi, & Sadri, 2020). For instance, with the spread of COVID-19 over the past few years, researchers have analyzed the sentiments of COVID-19-related tweets (Salmani et al., 2021; Vahdat-Nejad et al., 2022) and extracted the primary topics discussed (Azizi, Vahdat-Nejad, Hajiabadi, & Khosravi, 2021).

The incidence of wars and conflicts is among the occasions that prompt users to post their opinions on social networks. In this regard, users' views in various countries regarding recent events in

Afghanistan were analyzed from August to November 2021 (Aggarwal et al., 2022). For this purpose, the eight nations with the most tweets were investigated. One of the findings of this study was the high number of tweets containing negative terms such as *terrorist*, *attack*, *destruction*, and *violence*, as well as the fact that all eight investigated countries shared a similar perspective regarding the recent events in Afghanistan (Aggarwal et al., 2022). Likewise, a sentiment analysis (positive, neutral, negative, very positive, and very negative) was conducted on English and Turkish tweets related to the Syrian civil war in April 2017 (Öztürk & Ayvaz, 2018). The R-Sentiment package was used to analyze the sentiments of tweets written in English. Due to the substantial influx of Syrian refugees into Turkey, Turkish tweets were also analyzed. In the absence of a model for sentiment analysis of tweets written in Turkish, a sentiment lexicon of Turkish vocabulary was presented. Accordingly, there was a large number of positive tweets in Turkish and a large number of neutral, negative, and extremely negative tweets in English. Additionally, English tweets were more political and focused more on the legality of immigrants, whereas the Turkish-language tweets focused more on the war's details, as the conflict in proximity to the Turkish border was of interest to the Turkish community (Öztürk & Ayvaz, 2018). Similar sentiment analyses have been conducted for the ten countries with the highest volume of tweets regarding Syria's chemical warfare (Bashir et al., 2021). Overall, a greater proportion of tweets were negative, indicating that users had a negative view regarding the Syrian chemical attack (Bashir et al., 2021).

The Russia-Ukraine war is one of the most important events of 2022, about which countless tweets have been published. Recently, Chinese Weibo texts have been investigated to extract the main topics discussed by Chinese users about this conflict(Chen et al., 2022). We aim to continue that research by modeling the sentiment trends of various countries and clustering them. To the best of our knowledge, no study has been conducted to investigate and classify the sentiments trends of users from different countries regarding this war.

## 3- Proposed Method

Numerous tweets have been posted regarding the Russia-Ukraine war, which is one of the most significant events in 2022. To the best of our knowledge, no research has been conducted to analyze the sentiments of tweets related to this war and to examine and classify the sentiments of users from various countries. This research aims to collect and analyze tweets pertaining to the Ukrainian dispute in order to classify the sentiment trends of users from varying countries over time.

### 3-1- Tweet collection and processing

In March 2022, the war between Ukraine and Russia was the prominent topic of tweets. First, tweets related to the war in Ukraine are collected and separated by location. Afterward, positive and negative tweets are identified using sentiment analysis. The final step concerns clustering the time series of tweet sentiments.

We collected 140,000 English tweets associated with the Russia-Ukraine war using the keyword *Ukraine* during March 2022. The geotag is then applied to separate tweets from different countries. Thus, tweets with an unknown location are deleted.

The process of determining whether a tweet is positive, negative, or neutral is known as tweet sentiment analysis. A large share of sentiment analysis research is conducted on social network posts, such as those on Twitter, since these posts contain users' views and feelings. We utilize sentiment analysis to determine the trends of public sentiment regarding this war over time. To calculate the sentiment scores of tweets, we employed the promising language-based RoBERTa

(Liu et al., 2019) model. Recommended by Google, RoBERTa (Liu et al., 2019) is the optimized version of the language-based BERT (Devlin, Chang, Lee, & Toutanova, 2019) model. Identifying positive and negative tweets gives us insight into the direction of public opinion regarding the Ukraine conflict. The number of positive and negative tweets for each country (for which there are sufficient tweets) was then calculated weekly to yield a time series for each country. The analysis and clustering of time series pertaining to various nations unveiled intriguing information regarding public opinion.

### 3-2- Clustering countries' sentiments trends

Various methods have been presented for modeling and clustering users' opinions in social networks (Asadolahi, Akbari, Hesamian, & Arefi, 2021; G. Hesamian & M. Akbari, 2021; Hesamian & Akbari, 2018; G. Hesamian & M. G. Akbari, 2021; Hesamian & Akbari, 2022). This research draws on the neighborhood average method to cluster countries. Because the number of tweets in different countries varies, with the majority of these differences being substantial, the weekly data for each country is first normalized. Subsequently, the weekly frequency of the number of positive and negative tweets are computed for each country. Therefore, four positive and four negative frequencies corresponding to the first four weeks of the war's onset are calculated for each country and modeled as a vector with eight features. The neighborhood average method is then used to cluster the coefficients obtained from the proposed model for the nations. The distance between countries is calculated using the Euclidean meter, and the countries with a smaller distance are put in one cluster. The proposed model, which uses the support vector method (SVM), is elaborated on below.

First, we consider the following model for time data $\{Z_t: t = 1, 2, \dots, T$ (T = 8)$\}$:

$$Z_t = \sum_{j=1}^{p} W_j Z_{t-j} + W_0 = \vec{W}^T \vec{Z_t} + W_0 \qquad \vec{Z_t} = (Z_{t-1}, \ldots, Z_{t-p})$$

To estimate the vector $\vec{w}$ and $w_0$, we use the SVM as per figure1 and following equations:

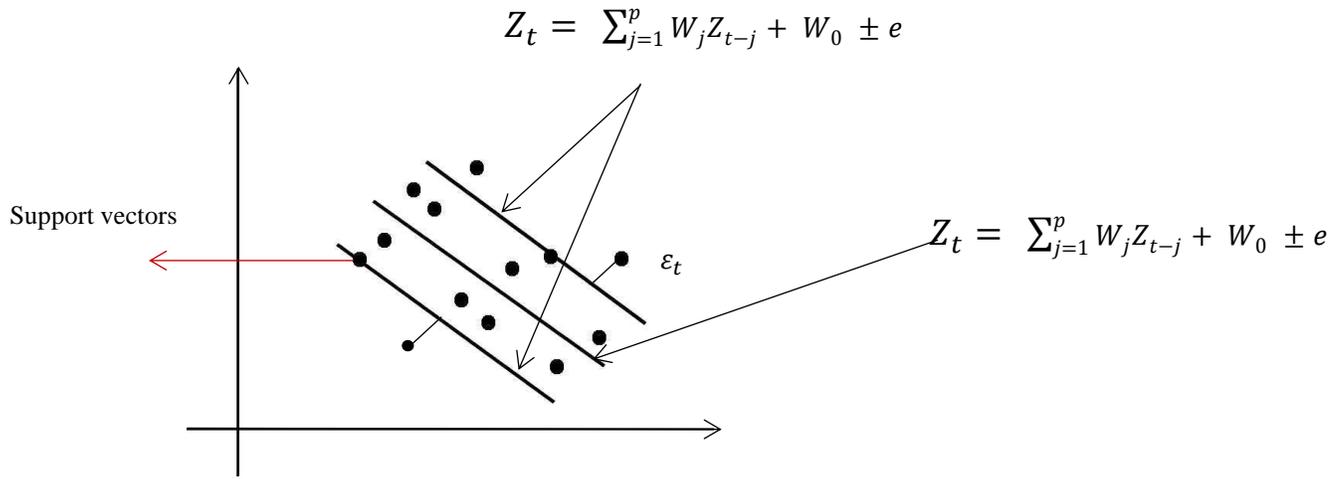

**Figure 1.** Using SVM for estimation

$$\min_{\vec{W}, W_0, \vec{\varepsilon}, \vec{\varepsilon}^*} J(\vec{W}, W_0, \vec{\varepsilon}, \vec{\varepsilon}^*) = \frac{1}{2} \vec{W}^T \vec{W} + c \sum_{t=p+1}^{T} (\varepsilon_t + \varepsilon_t^*) \qquad (1)$$

S.T

$$Z_t - \vec{W}^T \vec{Z_t} - W_0 \leq e + \varepsilon_t \qquad t = p+1, \ldots, T$$

$$\vec{W}^T \vec{Z_t} + W_0 - Z_t \leq e + \varepsilon_t^* \qquad t = p+1, \ldots, T$$

$$\varepsilon_t, \varepsilon_t^* \geq 0$$

Lagrange's equation for the above optimization problem with coefficients $\alpha_t, \alpha_t^*, \mu_t, \mu_t^*$ is as follows:

$$L(\vec{W}, W_0, \vec{\varepsilon}, \vec{\varepsilon}^*, \vec{\alpha}, \vec{\alpha}^*\mu, \vec{\mu}^*)$$

$$= J(\vec{W}, W_0, \vec{\varepsilon}, \vec{\varepsilon}^*)$$

$$-\sum_{t=p+1}^{T} \alpha_t \left(e + \vec{\varepsilon} - Z_t + \vec{W}^T \vec{Z_t} + W_0\right)$$

$$-\sum_{t=p+1}^{T} \alpha_t^* \left(e + \vec{\varepsilon}^* + Z_t - \vec{W}^T \vec{Z_t} - W_0\right) - \sum_{t=p+1}^{T} (\mu_t \varepsilon_t + \mu_t^* \varepsilon_t^*) \quad (2)$$

Where $L$ is used in Eq (2) to reach the optimal value $J$ in Eq (1), i.e.,

$$\max(\vec{\alpha}, \vec{\alpha}^*, \vec{\mu}, \vec{\mu}^*) \quad \min(\vec{W}, W_0, \vec{\varepsilon}, \vec{\varepsilon}^*) \quad L(\vec{W}, W_0, \vec{\varepsilon}, \vec{\varepsilon}^*, \vec{\alpha}, \vec{\alpha}^* \mu, \vec{\mu}^*)$$

We have:

$$\frac{\partial L}{\partial \vec{W}} = 0 \;\rightarrow\; \vec{W} = \sum_{t=p+1}^{T} (\alpha_t - \alpha_t^*) Z_t$$

$$\frac{\partial L}{\partial W_0} = 0 \;\rightarrow\; \sum_{t=p+1}^{T} (\alpha_t - \alpha_t^*) = 0 \quad (3)$$

$$\frac{\partial L}{\partial \varepsilon_t} = 0 \;\rightarrow\; c - \alpha_t - \mu_t = 0 \qquad \frac{\partial L}{\partial \varepsilon_t^*} = 0 \;\rightarrow\; c - \alpha_t^* - \mu_t^* = 0$$

By embedding Eq (3) in (2), the dual (1) is yielded as follows:

$max_{\vec{\alpha},\vec{\alpha}^*} J_D(\alpha, \alpha^*)$

$$= -\frac{1}{2} \vec{W}^T \vec{W}$$

$$- e \sum_{t=p+1}^{T} (\alpha_t + \alpha_t^*)$$

$$+ \sum_{t=p+1}^{T} Z_t(\alpha_t - \alpha_t^*)$$

$$= \sum_{t=p+1}^{T} \sum_{k=p+1}^{T} (\alpha_t - \alpha_t^*)(\alpha_k - \alpha_k^*) \vec{Z_t}^T \vec{Z_k}$$

$$- e \sum_{t=p+1}^{T} (\alpha_t + \alpha_t^*) + \sum_{t=p+1}^{T} Z_t(\alpha_t - \alpha_t^*)$$

S.T

$$\sum_{t=p+1}^{T} (\alpha_t - \alpha_t^*) = 0 \qquad 0 \leq \alpha_t, \alpha_t^* \leq c$$

The KKT method is used to obtain $W_0$ as follows:

$$\alpha_t \left( e + \varepsilon_t - Z_t + \vec{W}^T \vec{Z_t} + W_0 \right) = 0$$

$$\alpha_t^* \left( e + \varepsilon_t^* + Z_t - \vec{W}^T \vec{Z_t} - W_0 \right) = 0$$

As a result, with a few simple calculations, we have:

$$W_0 = Z_t - \vec{W}^T \vec{Z_t} - e \qquad \alpha_t \epsilon [0, c] \qquad \varepsilon_t = 0$$

$$W_0 = Z_t - \vec{W}^T \vec{Z_t} + e \qquad \alpha_t^* \epsilon [0, c] \qquad \varepsilon_t^* = 0$$

Therefore:

$$\widehat{W_0} = \frac{1}{|S|} \sum_{t \epsilon S} \left( Z_t - \vec{W}^T \vec{Z_t} - Sign(\alpha_t - \alpha_t^*)e \right)$$

$$S = \{t: 0 < \alpha_t - \alpha_t^* < c\}$$

If so, we will have:

$$\widehat{Z_t} = \sum_{k=p+1}^{T} (\hat{\alpha}_k - \hat{\alpha}_k^*) \vec{Z_k}^T \vec{Z_t} + \widehat{W_0}$$

Now, if we put $Z_t = \vec{W}^T \varphi(\vec{Z_t}) + W_0$ where $\varphi: R^p \to R^n$, we will similarly have:

$$\widehat{Z_t} = \sum_{k=p+1}^{T} (\alpha_k - \alpha_k^*) \varphi^T(\vec{Z_k}) \varphi(\vec{Z_t}) = \sum_{k=p+1}^{T} (\alpha_k - \alpha_k^*) K(\vec{Z_k}, \vec{Z_t}) \qquad (4)$$

in which $k$ is a kernel function. Instead of model (4), we consider the following model:

$$Z_t = \sum_{j=p+1}^{T} W_j K_h(\vec{Z_j}, \vec{Z_t}) + W_0$$

This model has many characteristics, including:

- The target equation is derived from the SVM method.
- Instead of using errors $\varepsilon_t, \varepsilon_t^*$, the loss function $\rho$ is used to estimate coefficients.
- The kernel function $K_h$ is used in dual equations in the model.
- There are fewer time twists when programming to obtain the values of the coefficients of the above equation.

- The smoothing constant can be calculated using the trial-and-error method or the generalized Wasserman cross-validation criterion.
- The optimal value of the smoothing constant $h$ in the loss function makes it possible to consider many outlier data and alterations in linear or non-linear $Z_t$ modes.

All $Z_p, \ldots, Z_{p+1}$ observations via function $\sum_{j=p+1}^{T} W_j K_h(\vec{Z}_j, \vec{Z}_t)$, which is the weighted sum of the neighborhood to the center $Z_t$ and radius $h$, are used to estimate $Z_t$. Besides, $h$ is the smoothing constant obtained by the GCV criterion. Moreover, we use the following objective equation to obtain $W_0$ and $\vec{W}$:

$$(\widehat{W}, W_0) = \min_{\vec{W}, W_0} \frac{\vec{W}^T W}{2} + c \sum_{t=p+1}^{T} \rho(Z_t - \sum_{j=p+1}^{T} W_j \, k \, (\vec{Z}_j, \vec{Z}_t))$$

, where $\rho$ is the loss function.

The mean least squared error ($\rho_{LS}$), which measures the distance between predicted and actual values, is one of the most well-known and widely used loss functions in the analysis and modeling of time-dependent data. Occasionally, the data is arranged such that the predicted values tend towards the outlier data and are so-called "crooked". In this case, the loss function mentioned above leads to problems when estimating parameters and predicting the response variable. Huber's loss function ($\rho_H$) is utilized to solve such a problem. In most modeling problems involving real-world data, we must determine whether predictions are certain or uncertain. Knowledge of the range of variations for predicted values is crucial for solving real-world issues. Using the quantile loss function ($\rho_Q$) has the property of providing an interval for the response variable rather than a specific value as a prediction. The forms of the aforementioned functions are summarized in Table 1.

**Table 1.** Loss functions

| methods | loss function |
|---|---|
| Least-Squares | $\rho_{LS}(e) = e^2$ |
| Huber | $\rho_H(e) = \begin{cases} \dfrac{1}{2}e^2 & for\ |e| \leq k \\ k|e| - \dfrac{1}{2}k^2 & for\ |e| > k \end{cases}$ |
| Quantile | $\rho_Q(e) = e(Q - I_{(e<0)}), \quad 0 \leq Q \leq 1$ |

**4-Evaluation**

The dataset contains 140,000 tweets related to the Russia-Ukraine war, collected during the first month of the war (March 2022) using the keyword *Ukraine*. The Python programming language was utilized for all implementations. Geotags were used to determine the tweets posted from each country, and sentiment analysis was performed on each country's tweets. At this stage, retweets were deleted to avoid duplication. Figure 2 depicts the frequency of tweets with positive and negative sentiments during the initial four weeks of the war per country. In all countries, the number of negative tweets exceeds the number of positive tweets, indicating that users have a negative view of the conflict in Ukraine. In addition, the proportion of countries that shared sufficient tweets about the conflict is larger in Europe, with European states accounting for 50 percent of the countries. Asian nations came in second place and made up nearly 30% of the nations. Switzerland, Singapore, Portugal, Ukraine, Spain, Italy, Austria, and Turkey have a higher negative to positive ratio of tweets, indicating a more negative attitude towards the events of the

Russia-Ukraine war. In contrast, the positive to negative tweet ratio is higher in Belgium, Denmark, China, Argentina, the Philippines, and Sweden. This ratio indicates that the citizens' opinions of these nations were less negative, in the first month of the conflict.

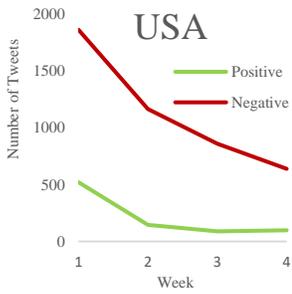 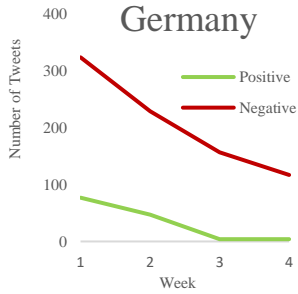 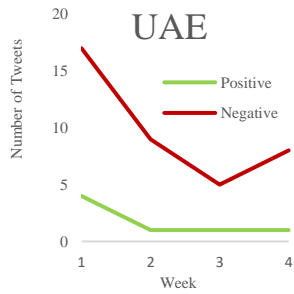 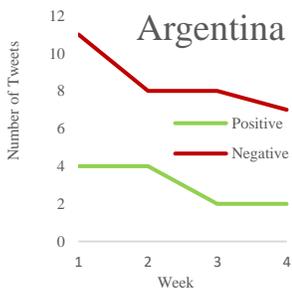
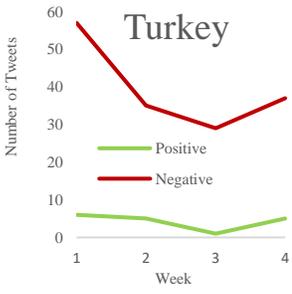 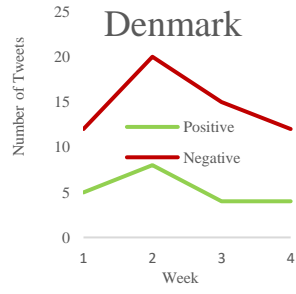 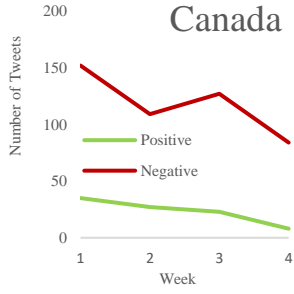 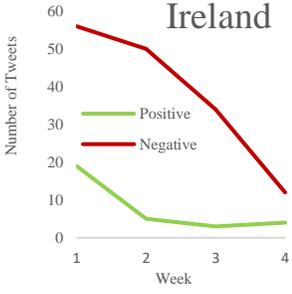
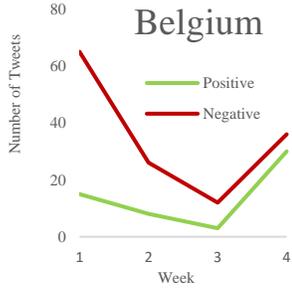 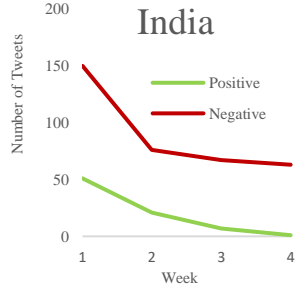 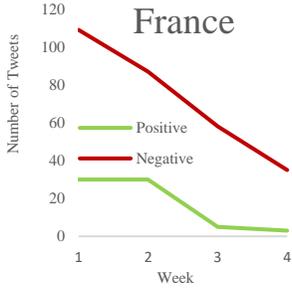 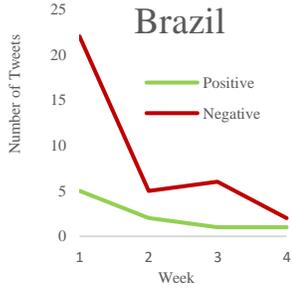
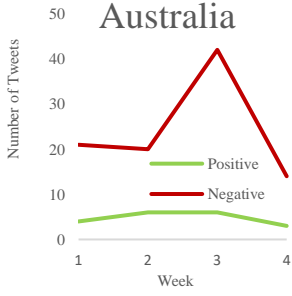 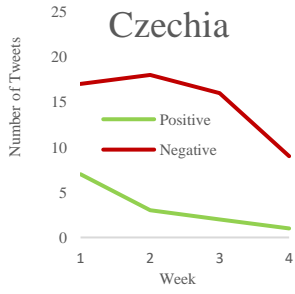 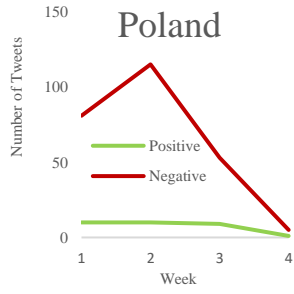 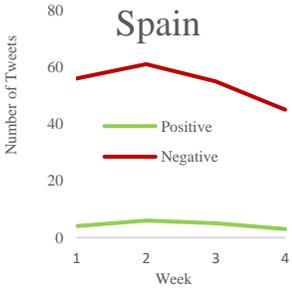

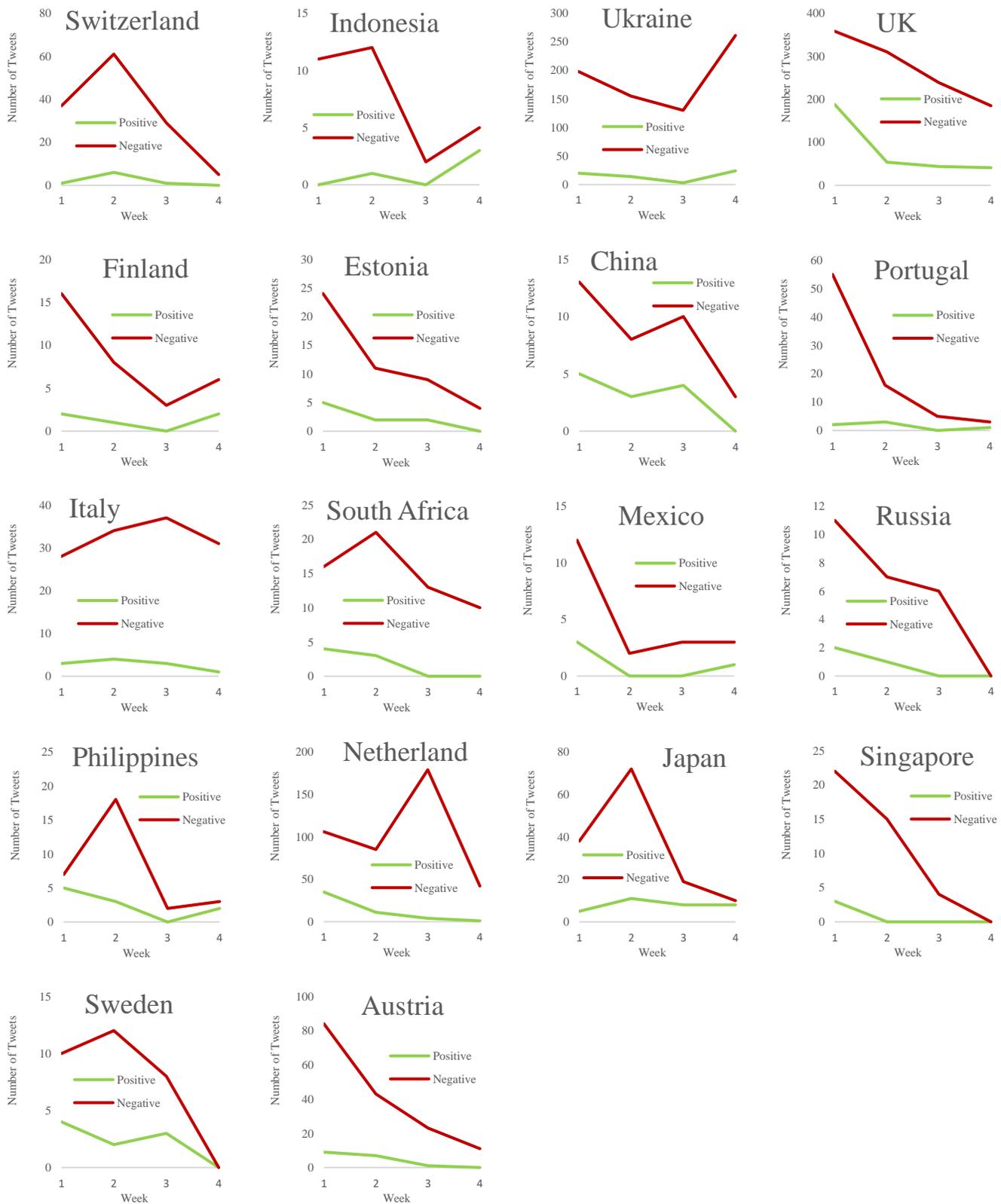

**Figure 2.** Frequency of positive and negative tweets in the first four months of the war

## 1-4- Clustering time series of countries

The proposed clustering model is applied to the 34 countries with the highest number of tweets. These 34 nations are ultimately classified into five clusters. Figure 3 illustrates the clustering of the countries. The greatest number of countries are located in clusters 1 and 5. Cluster 1 consists of the United States, Canada, England, India, and the majority of Western European nations. Except for India, all other states have fully backed Ukraine during the war. Cluster 2 only contains Ukraine. Given that this nation has been attacked, it is understandable that their attitude regarding this war differs from other nations.

Although the Japanese government sanctioned Russia, the model obtained for Japan is surprisingly unlike any other country. Cluster 4 is composed of Australia, Italy, and Spain. Ultimately, Cluster 5 comprises the United Arab Emirates, Estonia, Russia, Singapore, Finland, Portugal, Brazil, Argentina, Mexico, China, Denmark, Belgium, the Czech Republic, Poland, South Africa, Switzerland, the Philippines, and Sweden, all of which held relatively similar opinions with Russia. Most of these nations are in Asia, southern and central Africa, eastern Europe, and Scandinavia. When interpreting the clustering, it is important to consider the following two points, which also reflect the limitations of the work:

- The clustering of countries was performed according to the users' tweets in the first month of the war. Naturally, the type of clustering may change as the war continued and other events occurred.

- The dynamic trend of the time series of positive and negative tweets was clustered. As a result, the time parameter and the importance of the topic over time have been somehow involved.

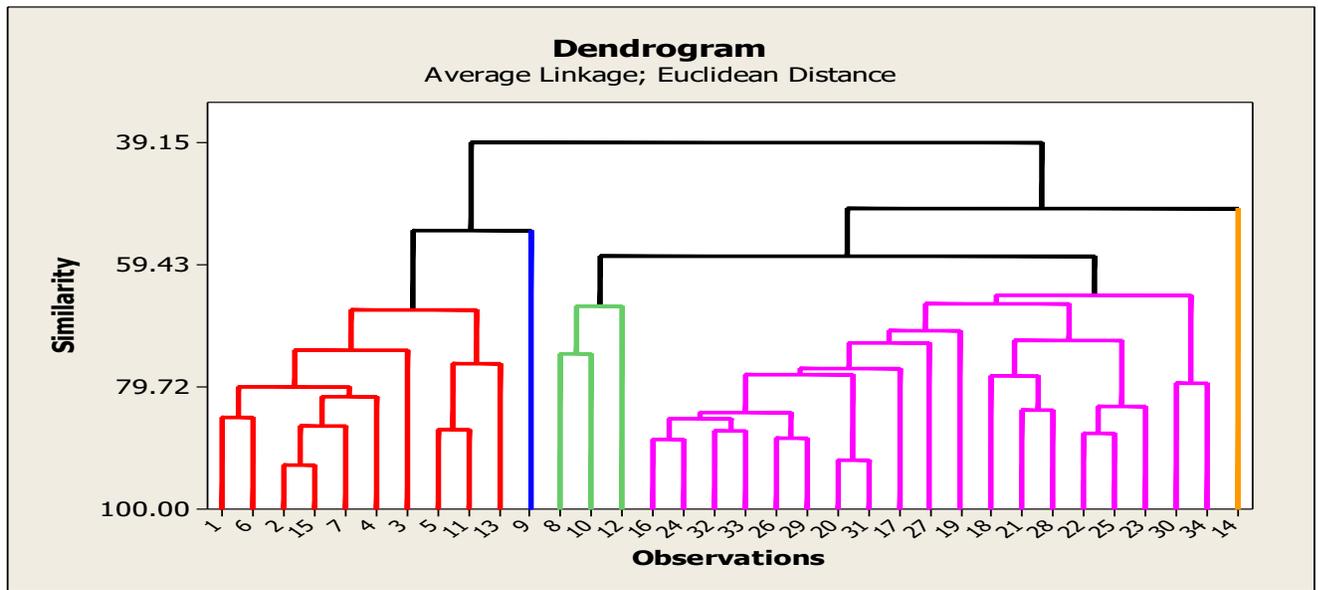
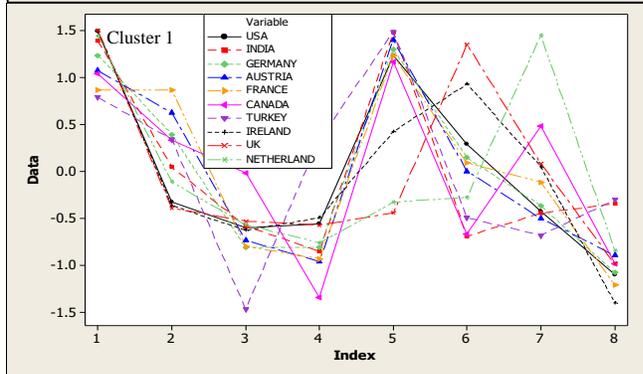
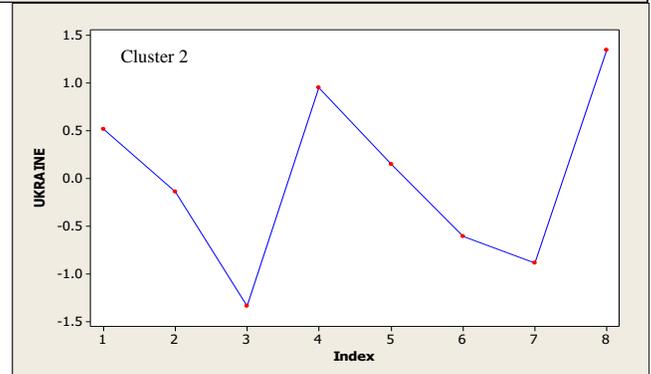
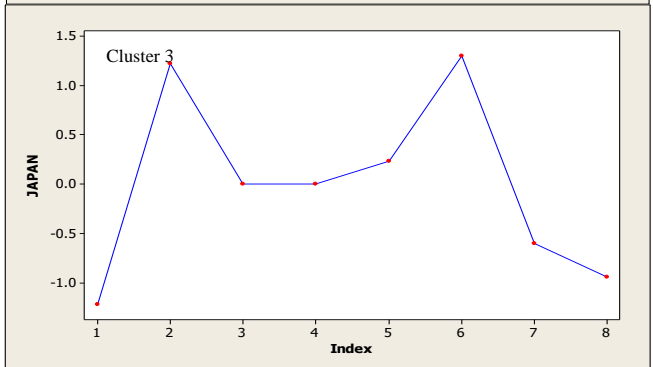
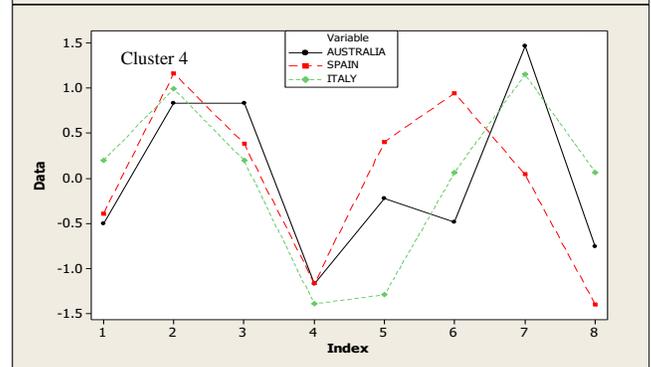
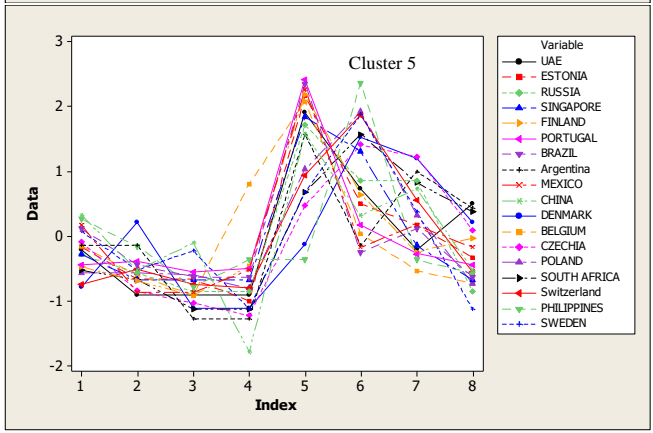

**Figure 3.** Clustering of 34 countries based on neighborhood average values

## 5. Conclusion

This article examines user sentiments regarding the Russia-Ukraine conflict during the first month of the conflict. For this purpose, 140,000 related English tweets were collected through the keyword Ukraine in March 2022. Then, the location of each tweet was determined based on its geotag. RoBERTa, a language-based model with superior performance to similar models, was used to analyze the sentiments of tweets. Afterward, the weekly time series of frequencies of positive and negative tweets from countries with a sufficient number of tweets were analyzed. The analysis of these time series yielded significant insight into users' perspectives regarding the Russia-Ukraine conflict. Because the number of negative tweets was greater than the number of positive tweets in all countries, it is safe to conclude that most users have a negative view of this war.

Furthermore, the trend of positive and negative tweets of the countries was clustered, which accounted for the topic's importance over time. The findings of this study allow us to draw attention to the similarity of views held by users in the United States, Canada, and Western Europe during the first month of the war, as well as the similarity of opinions held by users based in Eastern Europe, South America, Asia, and Scandinavia. In addition, users' views in Ukraine and Japan were distinct and unlike those of any other nation.

One of the limitations of this article is that it does not consider tweets in languages other than English, as users in many countries do not speak English fluently and publish tweets in other languages. Additionally, we have only used Twitter for polling purposes, and other social networks were omitted. Lastly, only tweets containing the keyword *Ukraine* were collected; therefore, all relevant tweets might not have been captured.

In future work, the coefficients derived from the proposed model for each country can be clustered based on various methods, and the results can be compared. Besides, the effects of this war on international relations, as well as economic, political, and other issues, can be analyzed from the perspective of users of social networks.

**Reference**

Abbasi-Moud, Z., Vahdat-Nejad, H., & Mansoor, W. (2019). *Detecting Tourist's Preferences by Sentiment Analysis in Smart Cities.* Paper presented at the Global Conference on Internet of Things, Dubai, United Arab Emirates, United Arab Emirates.
Aggarwal, S., Khan, S. S., & Kakkar, M. (2022). *Analyzing opinion of different countries for recent events in Afghanistan using text mining.* Paper presented at the 12th International Conference on Cloud Computing, Data Science & Engineering India.
Ansari, M. Z., Aziz, M., Siddiqui, M., Mehra, H., & Singh, K. (2020). *Analysis of political sentiment orientations on twitter.* Paper presented at the Computational Intelligence and Data Science, India.
Asadolahi, M., Akbari, M. G., Hesamian, G., & Arefi, M. (2021). A robust support vector regression with exact predictors and fuzzy responses. *International Journal of Approximate Reasoning, 132*(1), 206-225.
Asani, E., Vahdatnejad, H., Hosseinabadi, S., & Sadri, J. (2020). *Extracting User's Food Preferences by Sentiment Analysis.* Paper presented at the 8th Iranian Joint Congress on Fuzzy and intelligent Systems, Mashhad, Iran, Iran.
Azizi, F., Vahdat-Nejad, H., Hajiabadi, H., & Khosravi, M. H. (2021). *Extracting Major Topics of COVID-19 Related Tweets.* Paper presented at the International Conference on Computer Engineering and Knowledge, Iran.
Bashir, S., Bano, S., Shueb, S., Gul, S., Mir, A. A., Ashraf, R., & Noor, N. (2021). Twitter chirps for Syrian people: Sentiment analysis of tweets related to Syria Chemical Attack. *International Journal of Disaster Risk Reduction, 62*, 102397.
Briskilal, J., & Subalalitha, C. (2022). An ensemble model for classifying idioms and literal texts using BERT and RoBERTa. *Information Processing & Management, 59*(1), 102756.
Chen, B., Wang, X., Zhang, W., Chen, T., Sun, C., Wang, Z., & Wang, F.-Y. (2022). Public Opinion Dynamics in Cyberspace on Russia-Ukraine War: A Case Analysis With Chinese Weibo. *IEEE Transactions on Computational Social Systems, 9*(3), 948 - 958.
Corbett, J., & Savarimuthu, B. T. R. (2022). From tweets to insights: A social media analysis of the emotion discourse of sustainable energy in the United States. *Energy Research & Social Science, 89*, 102515.
Crannell, W. C., Clark, E., Jones, C., James, T. A., & Moore, J. (2016). A pattern-matched Twitter analysis of US cancer-patient sentiments. *journal of surgical research, 206*, 536-542.
Devlin, J., Chang, M.-W., Lee, K., & Toutanova, K. (2019). *BERT: Pre-training of Deep Bidirectional Transformers for Language Understanding.* Paper presented at the North American Chapter of the Association for Computational Linguistics: Human Language Technologies, USA.
Eisenstein, J. (2018). *Natural language processing*
Hesamian, G., & Akbari, M. (2021). Support vector logistic regression model with exact predictors and fuzzy responses. *Journal of Ambient Intelligence and Humanized Computing, 1*(1), 1-12.
Hesamian, G., & Akbari, M. G. (2018). A Semiparametric Model for Time Series Based on Fuzzy Data. *IEEE Transactions on Fuzzy Systems, 26*(5), 2953-2966.
Hesamian, G., & Akbari, M. G. (2021). A non-parametric model for fuzzy forecasting time series data. *Computational and Applied Mathematics, 40*(4), 1-21.
Hesamian, G., & Akbari, M. G. (2022). A fuzzy quantile method for AR time series model based on triangular fuzzy random variables. *Computational and Applied Mathematics, 41*(3), 1-20.
Jamalian, M., Vahdat-Nejad, H., & Hajiabadi, H. (2022). Investigating the Impact of COVID-19 on Education by Social Network Mining. *arXiv preprint arXiv:2203.06584*.
Liu, Y., Ott, M., Goyal, N., Du, J., Joshi, M., Chen, D., . . . Stoyanov, V. (2019). Roberta: A robustly optimized bert pretraining approach. *arXiv preprint arXiv:1907.11692*.
Mathur, A., Kubde, P., & Vaidya, S. (2020). *Emotional analysis using twitter data during pandemic situation: COVID-19.* Paper presented at the Communication and Electronics Systems India.
Öztürk, N., & Ayvaz, S. (2018). Sentiment analysis on Twitter: A text mining approach to the Syrian refugee crisis. *Telematics and Informatics, 35*, 136-147.
Salmani, F., Vahdat-Nejad, H., & Hajiabadi, H. (2021). *Analyzing the Impact of COVID-19 on Economy from the Perspective of User's Reviews.* Paper presented at the International Conference on Computer Engineering and Knowledge, Iran.
Sul, H. K., Dennis, A. R., & Yuan, L. (2017). Trading on twitter: Using social media sentiment to predict stock returns. *Decision Sciences, 48*, 454-488.

Vahdat-Nejad, H., Salmani, F., Hajiabadi, M., Azizi, F., Abbasi, S., Jamalian, M., . . . Hajiabadi, H. (2022). Extracting Feelings of People Regarding COVID-19 by Social Network Mining. *Journal of Information & Knowledge Management, 1*, 2240008.

Zad, S., Heidari, M., Jones, J. H., & Uzuner, O. (2021). *A survey on concept-level sentiment analysis techniques of textual data.* Paper presented at the IEEE World AI IoT Congress Virtual conference.